\documentclass{article}

 \usepackage[final]{neurips_2019}
\usepackage{natbib}
\usepackage[utf8]{inputenc} 
\usepackage[T1]{fontenc}    
\usepackage{hyperref}       
\usepackage{url}            
\usepackage{booktabs}       
\usepackage{amsfonts}       
\usepackage{nicefrac}       
\usepackage{microtype}      
\usepackage{graphicx}
\usepackage[caption=false]{subfig}

\newcommand{\rulesep}{\unskip\ \vrule\ }

\graphicspath{{images/}{examples/}{rw_examples/}}

\title{A general approach to bridge the reality-gap}

 \author{
   Michael R. Lomnitz\\ 
   In-Q-Tel Labs.\\
   Menlo Park, CA, 95025 \\
   \texttt{mllomnitz@gmail.com} \\
   \And
   Zigfried Hampel-Arias \\
   In-Q-Tel Labs.\\
  Menlo Park, CA, 95025 \\
   \And
   Nina Lopatina \\
   In-Q-Tel Labs.\\
   Menlo Park, CA, 95025 \\
   \And
   Felipe A. Mejia\\
   In-Q-Tel Labs.\\
   Menlo Park, CA, 95025\\
 }

\begin{document}

\maketitle

\begin{abstract}
Employing machine learning models in the real world requires collecting large amounts of data, which is both time consuming and costly to collect. A common approach to circumvent this is to leverage existing, similar data-sets with large amounts of labelled data. However, models trained on these canonical distributions do not readily transfer to real-world ones. Domain adaptation and transfer learning are often used to breach this “reality gap”, though both require a substantial amount of real-world data.  In this paper we discuss a more general approach: we propose learning a general transformation to bring arbitrary images towards a canonical distribution where we can naively apply the trained machine learning models.  This transformation is trained in an unsupervised regime, leveraging data augmentation to generate off-canonical examples of images and training a Deep Learning model to recover their original counterpart.  We quantify the performance of this transformation using pre-trained ImageNet classifiers,  demonstrating that this procedure can recover half of the loss in performance on the distorted data-set.  We then validate the effectiveness of this approach on a series of pre-trained ImageNet models on a real world data set collected by printing and photographing images in different lighting conditions.
\end{abstract}

\section{Introduction}
  With the relatively recent advent of large, labelled data-sets and ever cheaper compute, Machine Learning (ML), and particularly Deep Learning (DL), has garnered increased popularity.  Due to its many, sometimes superhuman, successes in fields such as computer vision or natural language processing, it has found itself in an increasing number of applications. However, in order to be successfully used in real applications, these models must correctly infer on data from the real world.
  
  Yet, it is often costly or time consuming to collect data specific to real life applications in which a model will ultimately be employed.  A common approach is to identify an existing data-set that resembles the ultimate application, and use this proxy to train ML models. However, networks that perform well during training and inference on this canonical data usually struggle when directly applied to these real life scenarios, due to differences that exist in the underlying distribution of the two data sets. 
  
   There are several established approaches designed to leverage existing knowledge in a network and adapt it to the target domain using limited data.  For instance transfer learning \cite{TransferLearning} is employed if there is access to a sample of \textit{labelled} data, while adversarial domain adaptation \cite{DomainAdaptation} can be used to adjust the network using \textit{unlabelled} data samples.  These methods are generally successful if provided sufficient training data from the target domain, however, both approaches require subsequent fine-tuning of each model individually.
 
  A different approach that has been recently discussed \cite{sim2sim} involves using neural networks to construct a function, $G$, that can take real life examples and translate them into samples that resemble the canonical data that the trained models understand.  This approach has many attractive features: first, $G$ can be trained in an unsupervised manner using examples from the existing canonical data-set. Second, it is ultimately agnostic to the models whose performance we hope to improve. As such, once trained, the translation network can be used to improve the performance of any network trained on the canonical set during inference on real world data.
  
In this paper, we propose a similar procedure to improve image classification on real images, captured in varying perspectives and lighting conditions.  We construct a general, learned transformation function to correct for these distortions to leverage powerful pre-trained classifiers.  This transformation is trained using data augmentation and existing data-sets (ImageNet \cite{ILSVRC15}) in a purely unsupervised regime. Furthermore, the procedure is agnostic to the specific model used for the classification task, eliminating the need to fine-tune individual models.
  
\section{Related work}
The use of synthetic data for training and testing DL networks has gained in popularity in recent years.  Relatively cheap to collect and label, it can be readily used to generate theoretically infinite amounts of data, including extremely rare events. 

However, in order to make models trained on these synthetic data-sets robust enough to function in real world applications it is necessary to ensure that they capture the variability of real world data.  Many approaches to bridge the reality gap have focused on Domain Randomization (DR) \cite{domain_rand} to generate synthetic data with enough variation that the models trained on them can generalize and view real world as another variation.  These approaches have found reasonable success in many models trained purely on synthetic data \cite{domain_rand, nvidia_det, flight_net}.

Nevertheless, DR has it's limitations.  Not only does it require large training times for the models to generalize beyond the randomized input and capture the important features, this makes the task potentially harder than is necessary.  The network now has to model the arbitrary randomization as well as capture the dynamics of the particular task.  Finally, each individual model has to be trained individually in this DR regime.

In this paper we take a different approach. Instead of training the DL model to be robust against this randomization, we construct a transformation that can revert the randomization and return images to a canonical form that any model, trained on canonical data, can understand.  Such a transformation is essentially agnostic to the ultimate application, and as such it can be used as a pre-processing step for any other downstream ML model. 

\subsection{Sim-to-real via sim-to-sim}
The authors in \cite{sim2sim} introduced a novel approach to cross the gap between data in real world applications and simulated environments during training of reinforcement learning (RL) grasping agent.  In robotic grasping, collection of real world data is not only expensive, but also cumbersome and time consuming.  And, as the availability of affordable cloud computing services increases, it has become increasingly attractive to train this task using simulated data.  With this, however, comes the issue of transferring the agent trained on the simulations to scenes from the real world, and the corresponding significant shift in the underlying distributions.

In their approach, they first trained the RL agent to grasp objects in a canonical simulation.  They then leveraged data augmentation and randomized the textures in these simulated environments and trained a Deep Learning (DL) network to transform them back into their canonical counterparts using an unsupervised image conditioned Generative Adversarial Network (GAN).
  
The authors then showed that this same transformation could be applied to real scenes, converting them into an equivalent canonical view via the same sim-to-sim transformation.  This allowed the grasping agent to vastly outperform (by a factor of 2 ) other approaches that did not require real grasps for training.  Finally, the authors fine tuned the model by optimizing the agent using a small set of online grasps. This attained performance akin to the state of the art, obtained by training using 580,000 offline, real grasps and 5,000 real, online grasps.

\subsection{Spatial Transformation Network}
Due to their relatively low number of parameters and ability to capture spatial invariance, Convolutional Neural Networks (CNN) have had great success in recent years in a broad set of applications such as classification, localization and segmentation.  However, due to the relatively small spatial scale of CNN kernels, achieving true spatial invariance requires a deep hierarchy of layers whose intermediate features are not invariant to large scale transformations.  Furthermore, the convolutional kernel does not capture the rotational invariance, requiring large amounts of data augmentation to train models to recognize familiar objects. 

The Spatial Transformer Network (STN) \cite{STN} was first introduced in an attempt to provide neural networks the capability to perform spatial transformations conditioned on individual samples, and whose behavior is learned during training for the designated task using standard back-propagation. Unlike the convolutional unit in CNNs, the spatial transformation module does not have a fixed receptive field, instead identifying the relevant sections or regions in images and then performing the transformations on the entire feature map.  These include translations as well as many transformations that can not be captured by CNNs, such as: rotations, scaling, cropping, as well as non-rigid transformations such as projections or skews.

The spatial transformation module consists of three separate tasks: first, a localization network identifies the relevant sections of the feature maps and the necessary transformation to be applied. Next, a grid generator identifies the positions in the input map from which to sample to produce the sampled output. Finally, the points in the sampling grid are sampled from the input map via a sampler to produce the output map.

\begin{figure}[h] \label{fig:STN}
  \centering
\subfloat[][]{
	\includegraphics[width=0.45\textwidth]{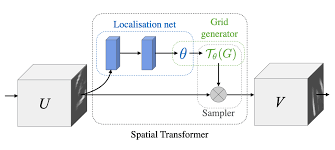}}
	\qquad
	\qquad
\subfloat[][]{
	\includegraphics[width=0.2\textwidth]{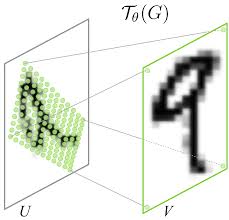}}

  \caption{a) Schematic figure taken from \cite{STN} illustrating the STN architecture consisting of localization network, grid generator and sampler. b) Example of applying the parametrized grid to an input image.}
\end{figure}

It is interesting to note that the STN module can act as a form of self attention, by focusing on the most impactful regions of a feature map that reduce the overall cost function for a given task. Multiple STN modules can be added in a CNN at different depths to produce transformations on more abstract features, or can be added in parallel to simultaneously locate multiple objects. Furthermore, the STN module is computationally very fast, and has minimal impact on the training time.

\section{Pipeline for Simulating real-world response}

In this section we discuss using Spatial Transformation Networks together with Adversarial Auto-Encoders to build such a \textit{real-to-canonical} transformation for models trained on ImageNet. The objective is to simulate the distortions that occur when images are captured using different cameras, in different lighting conditions, and off-center and slightly skewed.
   
    \subsection{Data generation: Distorted ImageNet}
  In order to learn the transformation $G$, we need to generate pairs of examples capturing data from the canonical distribution and its counterpart representing the real-life observation.  We use images from the ImageNet Large Scale Visual Recognition Challenge (ILSVR) 2012 \cite{ILSVRC15} data-set to test our approach by simulating the effect of capturing real-life images by applying a series of randomized distortions.
  
  The canonical (target) images, $y$, follow the standard evaluation pre-processing steps for the ILSVRC challenge in order to use available, pre-trained networks for later evaluation. The raw images are normalized using the mean and standard deviation of the entire training set,  their shorter axis is then re-scaled to 256 pixels and finally the image is center-cropped to produce 224x224 pixel inputs.
  
  For the distorted counter parts, $x$, we follow the same normalization and again re-scale such that the shorter axis measures 256 pixels and center-crop to 256x256 pixels.  We then apply random affine transformations allowing for rotations (\textbf{R} $\in[-5^{\circ}, 5^{\circ}]$), translations (\textbf{T} $\in[0.05, 0.05]$), scaling (\textbf{S} $\in[0.8, 1.1]$) and shear (\textbf{W}=5).  We also apply brightness, saturation, contrast and hue jitter by factors 0.25, 0.25, 0.25 and 0.1, respectively.  These random transformations are generated for each mini-batch at training time, such that the image-transformation pairs will vary over the different epochs.
  
\subsection{Model architecture}

\begin{figure}[h] \label{fig:STN_UNET}
  \centering
  \includegraphics[width = 0.95\textwidth]{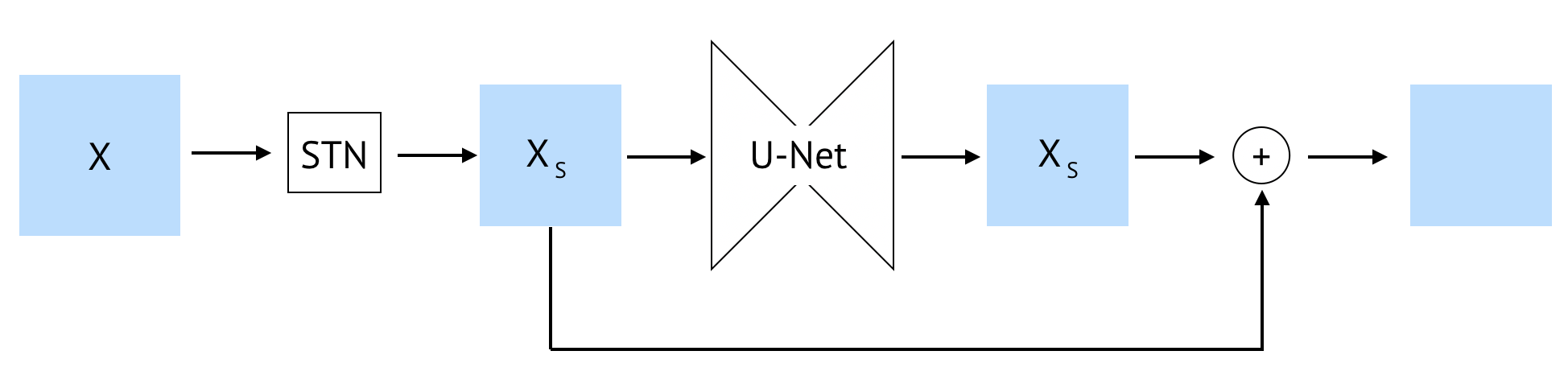}
  \caption{Schematic diagram illustrating the procedure to correct distorted images in the proposed model.}
\end{figure}

  We propose a two step procedure to correct the distortions in the input images $x$, illustrated in figure \ref{fig:STN_UNET}.  The images are first passed through a STN which uses the localization network, grid generator, and sampler to correct affine transformation component of the distortions in the input. The output of the STN is center-cropped to size 224x224 producing an intermediate step $x_s$.  The center-cropping at this stage serves dual purpose: not only does it provide the standard input for the pre-trained ImageNet models, but it also accounts for the majority of situations where part of the original image is lost due to random re-scaling and translations.
  The intermediate step $x_s$ is then passed through a U-Net auto-encoder, as described in \cite{UNET}.  We selected this particular architecture given that U-Net's have shown great success as auto-encoders on a variety of different tasks \cite{UNET, UNet-sharpening, 3D-Unet}.  By leveraging skip connections between the encoder and decoder halves, they are capable of producing quality reconstructions with relatively few parameters.  Nevertheless, the specific architecture for the auto-encoder is an open question and could be investigated in further work.
  
  The output of this second network is then added to $x_s$ to correct for the remaining color jitter in $x_s$ to produce the prediction $\hat{y}$.  We empirically observed that correcting for the hue change is a particularly challenging task. In order to facilitate the task, and following the reasoning laid out in \cite{CoordConv}, we provide the U-Net with the standard deviation for the three RGB color channels by concatenating them to the input images. Thus, the input to the U-Net is a tensor with dimensions (mini-batch $\times6\times224\times224$).
  
  \subsection{Training method}
    Our aim is to generate the transformation function $G(x) = y$ which corrects the randomly transformed images back to their canonical form. Assuming that the random transformations applied fully capture the variance observed in real life scenarios, this same transformation, $G$, will translate real-life examples back to the same canonical distribution.
    
  The model is trained using the mean squared error between the target image $y$ and the output of the STN-U-Net $\hat{y}$ ($\textrm{MSE}(y, \hat{y})$). It is well known that $\textrm{MSE}$ loss in auto-encoders can often lead to blurry generated images, given that it focuses on average differences between the target and prediction. To encourage the network to produce sharper, high fidelity reconstructions we include an adversarial loss from a discriminator trained simultaneously in a GAN framework.  The discriminator network consists of three CNN blocks (consisting of a 2D convolution followed by ReLU activation and batch norm) followed by a fully connected layer for the classification (real vs. fake), resulting in the following loss:
  
  \begin{eqnarray}
  \mathcal{L}_{eq}& = & \min_{G}\max_{D} \mathbb{E}_{y\in p_{c}}\left[\log D(y)\right] + \lambda\mathbb{E}_{x\in p_{r}}\left[\log (1-D(G(x))\right] \nonumber \\
  			&&	+ \mathbb{E}_{x, y \in p_{c}, p_{r}}\left[\textrm{MSE}(y, G(x))\right]
  \end{eqnarray}
  
   Finally, the contribution to the loss from the discriminator was weighted by a factor $\lambda = 1/4$. This value was selected through brief experimentation showing that it yielded desired results: it produced sharper images while allowing the network to focus on correcting the hue of the image.  

There are several key advantages to this procedure:
\begin{itemize}
\item Both $y$ and $\hat{y}$ come from the same distribution, and, as such, the problem is wholly unsupervised. 
\item The problem as posed is relatively simple since the STN-U-Net only needs to learn how to correct the images, and not the underlying distribution in ImageNet.
\item The transformations applied are randomized during each iteration, hence, the likelihood of the network seeing the same image/transformation pair is remote. Each epoch over the data-set results in augmenting the data-set and not over-fitting.
\item As long as the variance in the transformations captures the variance we would potential encounter in the ultimate application of the model, the \textit{real-to-canonical} transformation should provide intelligible outputs.
\end{itemize}

\subsection{Validation}
\begin{figure}[h] \label{fig:rw_examples}
  \centering
\subfloat{
	\includegraphics[width=0.31\textwidth]{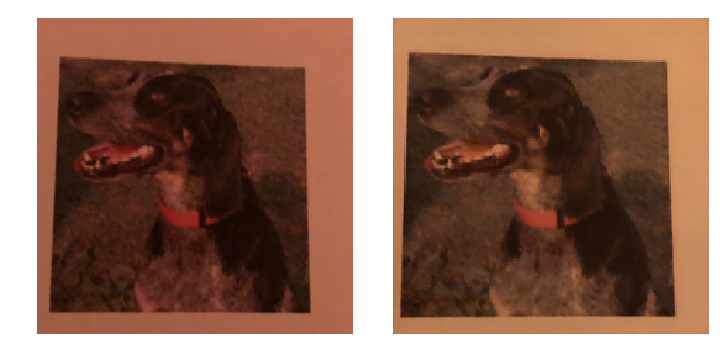}}
	\rulesep
\subfloat{
	\includegraphics[width=0.31\textwidth]{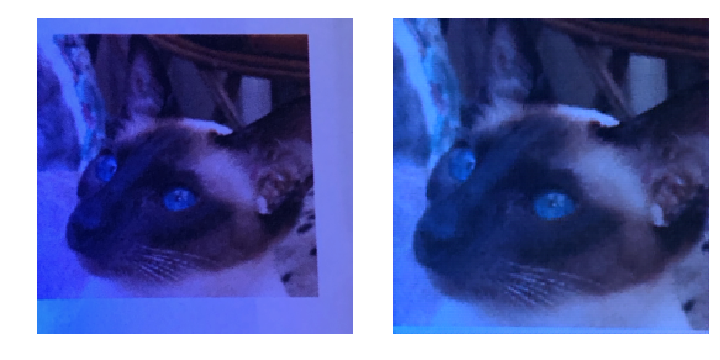}}
	\rulesep
\subfloat{
	\includegraphics[width=0.31\textwidth]{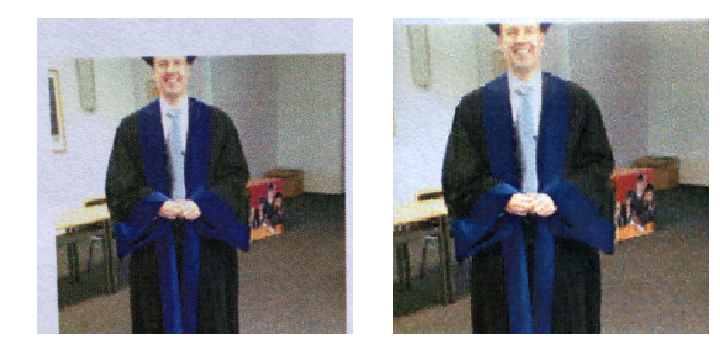}} \\
\subfloat{
	\includegraphics[width=0.31\textwidth]{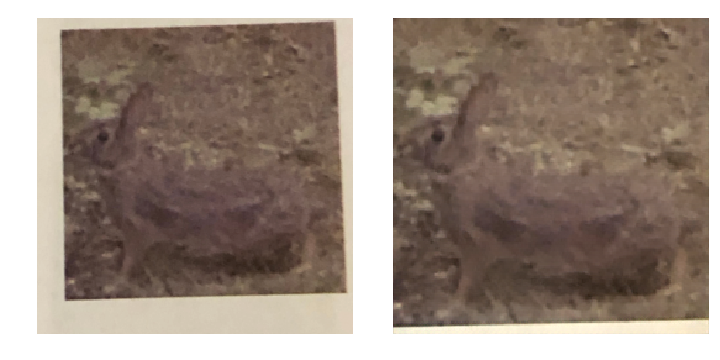}}
	\rulesep
\subfloat{
	\includegraphics[width=0.31\textwidth]{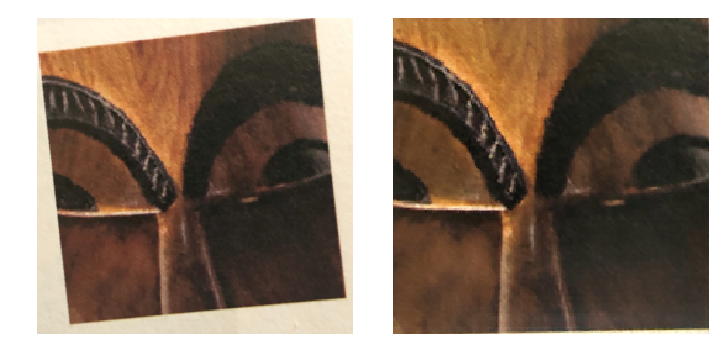}}
	\rulesep
\subfloat{
	\includegraphics[width=0.31\textwidth]{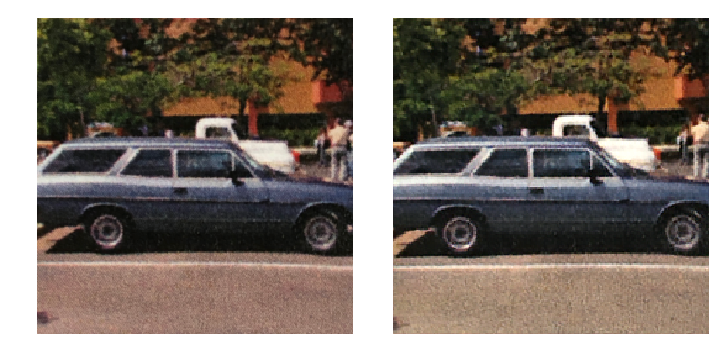}} \\
\subfloat{
	\includegraphics[width=0.31\textwidth]{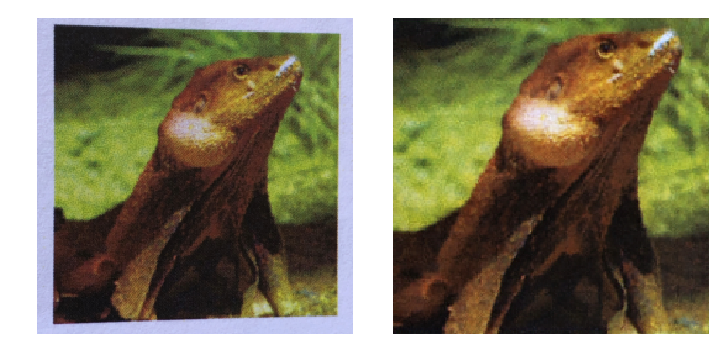}}
	\rulesep
\subfloat{
	\includegraphics[width=0.31\textwidth]{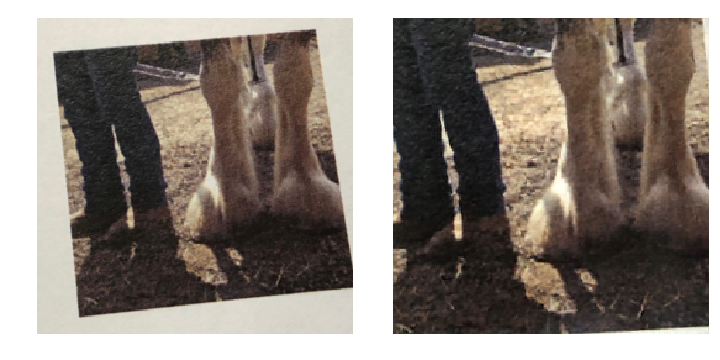}}
	\rulesep
\subfloat{
	\includegraphics[width=0.31\textwidth]{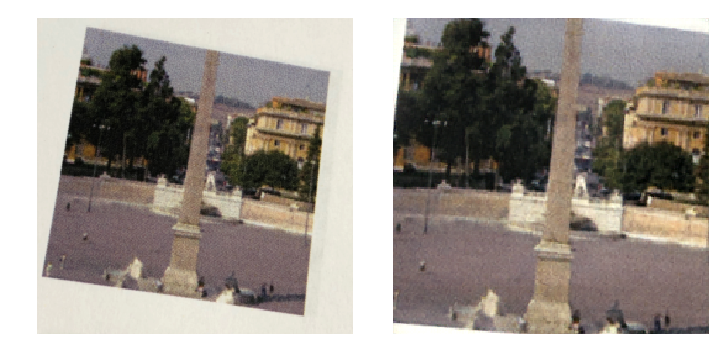}} \\
  \caption{Nine examples illustrating the canonical $y$ (left), the distorted $x$ (middle) and corrected (right) images in the experiment $\hat{y}$.}
\end{figure}

  To test the performance of the model on real world data, we printed a random sample of 400 1.5" x 1.5" inch and 300 5" x 5" images from ImageNet using a standard color laser printer. We then photographed the images under varying lighting conditions (hue and brightness) using a smartphone camera, and validated the same transformation $G$ in correcting the resulting distortions.  Figure \ref{fig:rw_examples}, in the following section, shows a sample of nine such images together with their associated, corrected pairs.
  
\section{Results}
\subsection{Distorted ImageNet}
\begin{figure}[h] \label{fig:examples}
  \centering
\subfloat{
	\includegraphics[width=0.45\textwidth]{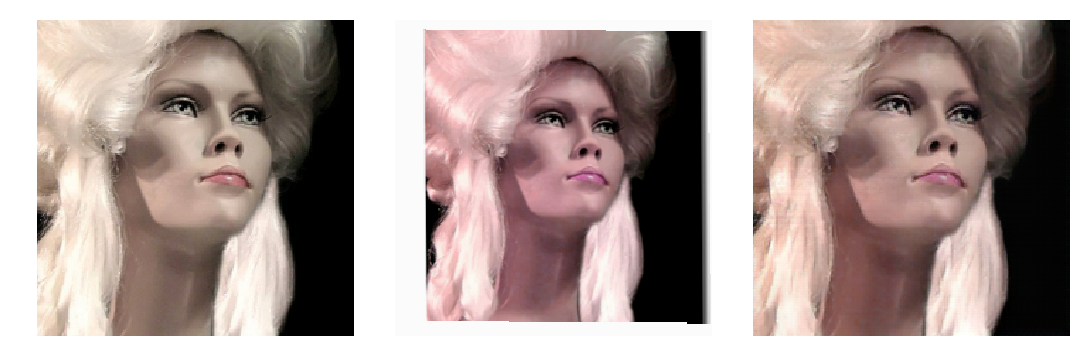}}
	\rulesep
\subfloat{
	\includegraphics[width=0.45\textwidth]{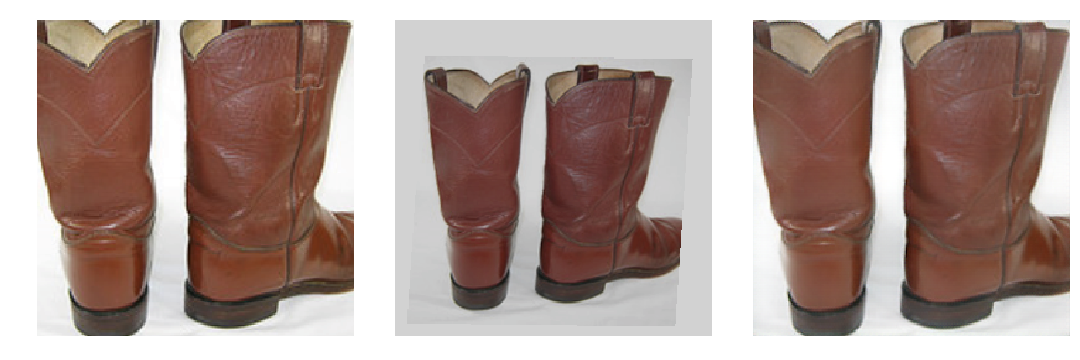}} \\
\subfloat{
	\includegraphics[width=0.45\textwidth]{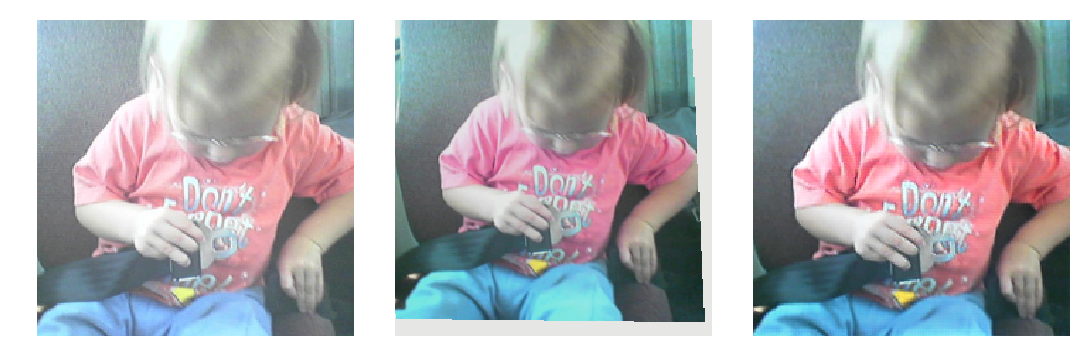}}
	\rulesep
\subfloat{
	\includegraphics[width=0.45\textwidth]{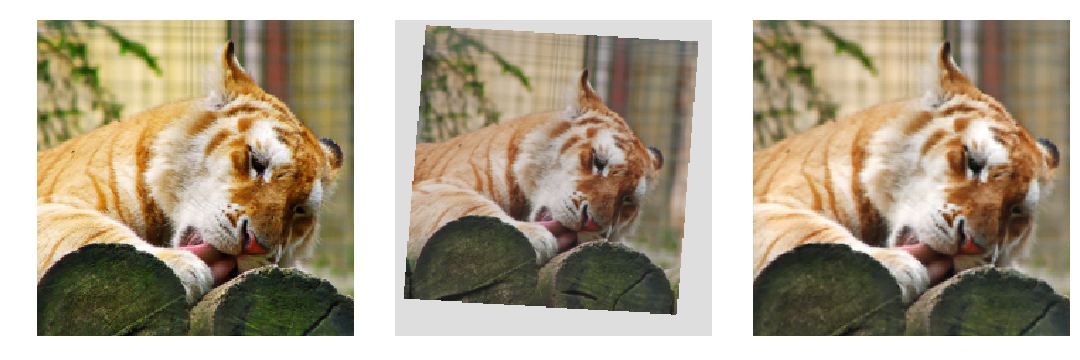}}	\\
\subfloat{
	\includegraphics[width=0.45\textwidth]{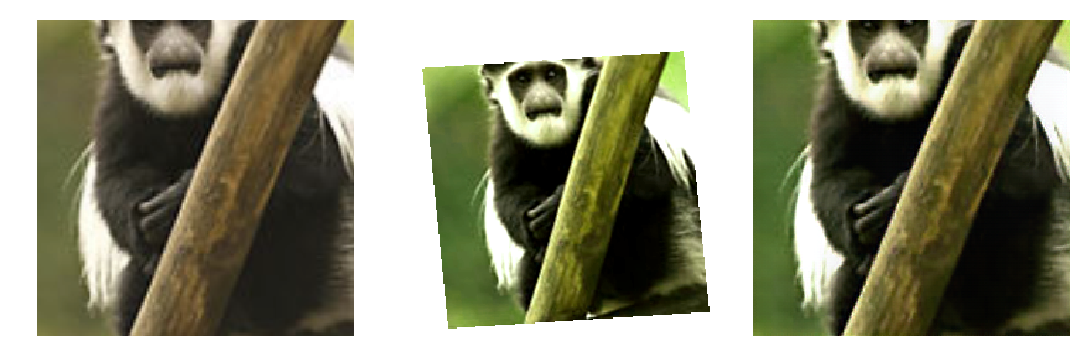}}
	\rulesep
\subfloat{
	\includegraphics[width=0.45\textwidth]{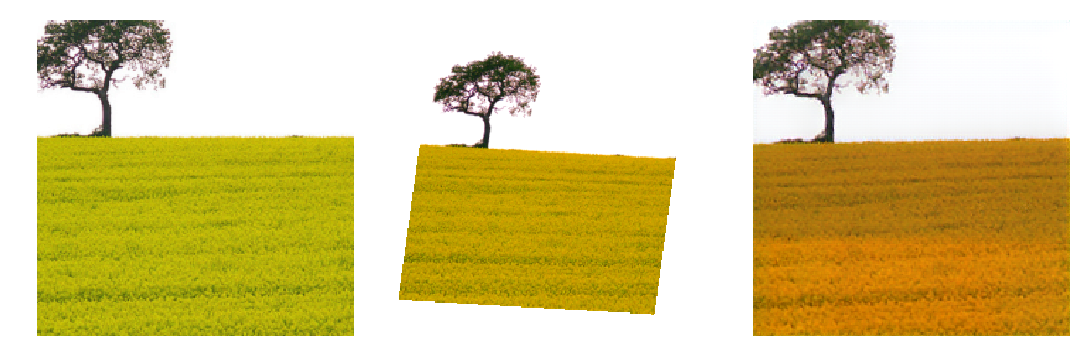}}
  \caption{Six examples illustrating the canonical $y$ (left), the distorted $x$ (middle) and corrected (right) images in the experiment $\hat{y}$.}
\end{figure}

  Figure \ref{fig:examples} shows six examples of the results of our experiments after training the transformation $G(x)$ for a single epoch over the entire 1.2M images. We experimented with 10 more iterations over the training data and found no significant improvement.  Generally, we see good agreement between the canonical and corrected images. In particular, the network excels at correcting the affine distortions, as is expected by the use of the STN.  On the other hand, hue jitter is particularly challenging for the network to correct. The bottom example on the right column is an interesting example where the network has over-corrected the hue, as the network expects images to be within a certain distribution determined by the standard deviation in each channel.
  
For a quantitative understanding of the performance of this \textit{real-to-canonical} transformation $G$ we compare the performance of a series of pre-trained ImageNet models \cite{resnet, squeezenet, vgg} on: a) the canonical images ($y$), b) the distorted images ($x$) and finally c) the images corrected via  the transform ($G(x) = \hat{y}$).  Table  \ref{table:tests_1_epoch} summarizes the top-1 accuracy (where the correct class is the top prediction from the classifier) obtained after a single epoch of training the STN-U-Net over the ILSVRC12 training set and evaluating over the validation set provided in the competition.  The last column, labelled "Improvement", indicates the difference between the classification on the raw, distorted images ($x$) and the classification on the corrected, distorted images ($G(x)$).

As expected, the performance of all of the models drops due to the distortions in the images relative to the canonical ImageNet data-set, although the better performing networks are more robust as they already include a good deal of data augmentation during training.  After correcting the distortions with the transformation $G$, about half to a third of the drop in performance in each model is recovered.  We note that the classifiers have not been fine-tuned on the output of the transformation $G$, illustrating the general purpose of this approach.
  
\begin{table}[h]
  \caption{Performance of 3 different pre-trained ImageNet models on the canonical ImageNet data ($y$), the distorted (affine and color jitter) counter parts ($x$) and the corrected images via the \textit{real-to-canonical} transformation $\hat{y}$.  The final column indicates the improvement obtained when including the transformation.}
  \label{table:tests_1_epoch}
  \centering
  \begin{tabular}{lcccc}
    \toprule
     & \multicolumn{4}{c}{Top-1 validation accuracy (\%)}                   \\
    \cmidrule(r){2-5}
    Model & Canonical & Distorted & Corrected & Improvement\\
     \midrule
    Resnet 50 \cite{resnet} & 76.13 &  67.55 & 71.26 & 3.71\\
    VGG 11 \cite{vgg}     & 69.02 & 57.87 & 62.69 & 4.82\\
    SqueezeNet 1.1 \cite{squeezenet} & 58.18 & 42.66 & 51.56 & 8.9 \\
    
    \bottomrule
  \end{tabular}
\end{table}

\subsection{Real World ImageNet}
 
 The following images illustrate how the transformation performed in generalizing to real life photographs.  The STN-UNet is capable of correcting many of the distortions in the images, though not as effectively as it did on our simulated data-set.  For instance, the top-left pair shows how the STN failed to localize and correct for the affine transformation and the top-middle pair could not correct for the extreme color hue distortion.

As was done in the previous section, we can use image classification as a proxy to validate the performance of the transformation $G$.  Table \ref{tab:tests_rw} shows the top-1 accuracy using the same pre-trained classifiers on the 1.5" x 1.5" and 5" x 5" real world images as well as their respective corrected versions.  In both cases the performance of the model drops, not only due to the distortions introduced due to lighting conditions and the angle, but also due to the printing of the images, as is evidenced by the stark drop in the smaller 1.5" x 1.5" images.  This suggests that our simulation pipeline could be further improved by adding random blurring to the images so that the function $G$ can account for these distortions as well. Nevertheless, the function trained on our simulated data-set can still correct for about 1-3\% of the loss in performance for all of the tested classifiers in both data-sets.

\begin{table}[h]
  \caption{Performance of 3 different pre-trained ImageNet models on real world images ($x$) and their counterparts corrected images via the \textit{real-to-canonical} transformation $G(x)=\hat{y}$.}
  \label{tab:tests_rw}
  \centering
  \begin{tabular}{lcccccc}
    \toprule
     & \multicolumn{6}{c}{Top-1 validation accuracy (\%)}  \\
     \cmidrule(r){2-7}
    Model & \multicolumn{3}{c}{1.5" x 1.5"} & \multicolumn{3}{c}{5" x 5"}                \\
     \cmidrule(r){2-4} \cmidrule(r){5-7}
     & Raw & Corrected & Diff. & Raw & Corrected & Diff.\\
     \midrule
    Resnet 50 \cite{resnet}             & 42.5 & 46.3 & 3.8 & 60.7 & 62.3 & 1.6\\
    VGG 11 \cite{vgg}                   & 26.5 & 29.3 & 2.8 & 48.3 & 51.7 & 3.4\\
    SqueezeNet 1.1 \cite{squeezenet}    & 21.5 & 23.5 & 2.0 & 36.0 & 38.0 & 2.0\\
    
    \bottomrule
  \end{tabular}
\end{table}

\section{Conclusions and future work}
  In this work we have developed a general scheme to bridge the "reality gap" for image data-sets relying on constructing a general \textit{real-to-sim} transformation $G$ designed to convert messy, real world data into a canonical distribution.  We demonstrated how models trained on this canonical set can all benefit from this transformation, potentially removing the need to perform domain adaptation or transfer learning on a model to model basis.
  
In our pipeline, this transformation was trained in a completely unsupervised regime relying on data augmentation to model potential deviations from the canonical distribution.  The performance of the learned transformation was validated by comparing the performance of a set of pre-trained models on the canonical set, the distorted set, and the corrected set, with significant improvements across the board. 

We also show that the transformation can be applied to images taken from real life under a variety of conditions. We collected two small validation sets by printing and photographing images from different angles and lighting conditions and applied the same transformation function $G$.  Again, the performance of the classification networks showed improvements in real life data after correcting with $G$. Though the results were less significant than in the distorted image data-set, this suggests an interesting avenue to develop a general, task independent scheme to bridge the gap between two different distributions.   Further improvement is under investigation, relying on more aggressive data augmentation in the manner of including larger distortions and account for a vital effect: resolution loss or blurring.


\bibliography{NIPS2019_STN_paper}

\end{document}